\documentclass{article}

\usepackage{microtype}
\usepackage{graphicx}
\usepackage{subcaption}
\usepackage{multirow}
\usepackage{booktabs} 
\usepackage{tabularx} 
\usepackage[table]{xcolor}
\usepackage{hyperref}

\usepackage[preprint]{icml2026}

\usepackage{amsmath}
\usepackage{amssymb}
\usepackage{mathtools}
\usepackage{amsthm}

\usepackage[capitalize,noabbrev]{cleveref}

\theoremstyle{plain}

\theoremstyle{definition}

\theoremstyle{remark}

\usepackage[textsize=tiny]{todonotes}

\icmltitlerunning{MIC: Maximizing Informational Capacity in Adaptive Representations via Isotropic Subspace Alignment}

\begin{document}

\twocolumn[
  \icmltitle{MIC: Maximizing Informational Capacity in Adaptive Representations via Isotropic Subspace Alignment}




  \begin{icmlauthorlist}
    \icmlauthor{Nguyen Hong Dang}{hust,vinshc}
    \icmlauthor{Nhi Ngoc-Yen Nguyen}{vinshc}
    \icmlauthor{Huy-Hieu Pham}{vinshc,vineng,cihs}
  \end{icmlauthorlist}

\icmlaffiliation{hust}{Hanoi University of Science and Technology, Hanoi, Vietnam}
\icmlaffiliation{vineng}{College of Engineering and Computer Science, VinUniversity, Hanoi, Vietnam}
\icmlaffiliation{vinshc}{VinUni-Illinois Smart Health Center, VinUniversity, Hanoi, Vietnam}
\icmlaffiliation{cihs}{Center for Innovations in Health Sciences, VinUniversity, Hanoi, Vietnam}

\icmlcorrespondingauthor{Huy-Hieu Pham}{hieu.ph@vinuni.edu.vn}

  \icmlkeywords{Matryoshka Representation Learning, Isotropic Subspace Alignment, Representation Collapse, Semantic Compression}

  \vskip 0.3in
]



\printAffiliationsAndNotice{}  

\begin{abstract}
Although multi-scales representation learning enables elastic-dimension embeddings, nested subspaces often suffer from dimensional redundancy and spectral collapse. To address this, we introduce \textbf{MIC}, a framework that optimizes the geometric landscape of multi-granular embeddings through isotropic subspace alignment. MIC employs Soft Collapse Regularization (SCR) to mitigate redundancy between prefix and residual subspaces via cross-correlation penalties, alongside Spectral Isotropy Regularization (SIR) to ensure hyper-spherical uniformity in low-dimensional prefixes. By unifying these strategies through a self-distillation objective, MIC generates semantically dense representations that maintain high discriminative power. Our experiments demonstrate that MIC significantly outperforms standard baselines, particularly in high-compression scenarios where maintaining informational capacity is most critical.
\end{abstract}

\section{Introduction}
\label{sec:intro}

The proliferation of large-scale dense vector representations has revolutionized modern Natural Language Processing \citep{pennington-etal-2014-glove, devlin2019bertpretrainingdeepbidirectional}, providing a unified interface for retrieval, clustering, and semantic search. However, the one-size-fits-all nature of standard embeddings creates a significant friction point in production: high-dimensional vectors offer superior accuracy but impose prohibitive storage and latency costs, while low-dimensional models often lack the expressive power required for complex reasoning \citep{wang2020minilmdeepselfattentiondistillation}. Matryoshka Representation Learning~\cite{NEURIPS2022_c32319f4} recently emerged as a versatile solution to this dilemma. By nesting multiple granularities within a single high-dimensional vector, MRL allows for adaptive inference-time truncation, enabling models to scale their computational footprint dynamically based on available resources.

Training robust Matryoshka models presents geometric challenges that standard objectives frequently overlook. While current MRL implementations \citep{NEURIPS2022_c32319f4, li2025ese} rely on multi-objective supervision across nested dimensions, this approach ensures basic functionality without guaranteeing optimality. Instead, models often suffer from representation collapse, where redundant or noisy signals degrade semantic discriminability and create a crowding effect that sacrifices valuable features. We argue that unlocking MRL’s full potential requires geometric conditioning to manage the representation space's informational capacity and spectral properties. When prefix subspaces are spectrally biased or highly correlated with residuals, the resulting informational bottleneck causes performance to plummet as dimensions decrease. Consequently, truly scalable embeddings require every dimension to contribute unique, non-redundant information while maintaining a uniform, isotropic distribution on the hyper-sphere.

Motivated by these geometric insights, we propose \textbf{MIC: Maximizing Informational Capacity in Adaptive Representations via Isotropic Subspace Alignment}, a framework that maximizes informational capacity through isotropic subspace alignment. During self-distillation, MIC employs SCR to minimize redundancy between prefix and residual subspaces and SIR to ensure a uniform, non-anisotropic embedding distribution. By optimizing these geometric properties, MIC ensures that interacting subspaces are structurally regularized for maximum efficiency and discriminative power. We also discuss the related works and background in Appendix~\ref{sec:rel_works}. Our main contributions are as follows:
\begin{itemize}
\item We introduce \textbf{MIC}, a unified framework that shifts the focus of Matryoshka training from simple usability to the maximization of informational capacity through geometric alignment.
\item We propose Soft Collapse Regularization, a thresholded correlation penalty that manages the informational boundary between nested subspaces, preventing redundancy without the rigid constraints of hard orthogonality. Furthermore, we develop Spectral Isotropy Regularization, which leverages hyper-spherical uniformity and spectral balancing to ensure well-conditioned representations across granularities.
\item Through extensive experiments on diverse tasks, we demonstrate that \textbf{MIC} consistently outperforms state-of-the-art MRL baselines. Most notably, our method maintains robust performance in extreme low-dimensional settings.
\end{itemize}

\section{Proposed Method}

\begin{table*}[h!]
\centering
\resizebox{\textwidth}{!}{%
\begin{tabular}{lc|cccc|cccc!{\vrule width 1.5pt}lc|cccc|cccc}
\toprule
\multirow{2}{*}{\textbf{Datasets}} & \multirow{2}{*}{\textbf{Dim.}} & \multicolumn{4}{c|}{\textbf{TinyBERT 6L}} & \multicolumn{4}{c!{\vrule width 1.5pt}}{\textbf{BERT}} & 
\multirow{2}{*}{\textbf{Datasets}} & \multirow{2}{*}{\textbf{Dim.}} & \multicolumn{4}{c|}{\textbf{TinyBERT 6L}} & \multicolumn{4}{c}{\textbf{BERT}} \\
\cmidrule(lr){3-6} \cmidrule(lr){7-10} \cmidrule(lr){13-16} \cmidrule(lr){17-20}
 & & \small{Unsup SimCSE} & \small{MRL} & \small{ESE} & \small{MIC} & \small{Unsup SimCSE} & \small{MRL} & \small{ESE} & \small{MIC} & 
 & & \small{Unsup SimCSE} & \small{MRL} & \small{ESE} & \small{MIC} & \small{Unsup SimCSE} & \small{MRL} & \small{ESE} & \small{MIC} \\
\midrule

\multirow{7}{*}{Banking77} 
 & 16 & 30.18 & 40.64 & 43.35 & \textbf{44.1} & 35.92 & 46.39 & 47.01 & \textbf{59.45} & \multirow{7}{*}{WIC} 
 & 16 & 58.21 & 57.35 & 57.5 & \textbf{58.64} & 56.57 & 59.35 & 58.35 & \textbf{61.64} \\
 & 32 & 56.23 & 62.59 & 63.78 & \textbf{65.7} & 54.23 & 64.9 & 63.63 & \textbf{75.71} & 
 & 32 & 60.07 & 59.07 & 58.64 & \textbf{60.18} & 60.85 & 61.5 & 60 & \textbf{62.07} \\
 & 64 & 65.48 & 73.58 & 73.96 & \textbf{77.52} & 67.78 & 76.84 & 76.24 & \textbf{83.05} & 
 & 64 & \textbf{62.07} & 60.85 & 59 & 60.89 & 62.85 & 63 & 62.14 & \textbf{63.71} \\
 & 128 & 75.32 & 80.31 & 81.53 & \textbf{83.49} & 75.43 & 83.49 & 82.94 & \textbf{86.57} & 
 & 128 & \textbf{62.78} & 62.5 & 60.14 & 62 & 63.78 & 63.93 & 61.71 & \textbf{64.65} \\
 & 256 & 84.94 & 85.04 & 85.76 & \textbf{86.52} & 85.84 & 86.45 & 86.16 & \textbf{88.11} & 
 & 256 & 62.64 & 62.21 & 60.07 & \textbf{62.71} & 64.91 & 64.78 & 63.07 & \textbf{64.92} \\
 & 512 & 87.42 & 85.63 & 86.92 & \textbf{87.93} & 88.34 & 87.85 & 88.51 & \textbf{89.43} & 
 & 512 & 62.46 & 62.69 & 59.92 & \textbf{62.78} & 64.58 & 64.14 & 63.07 & \textbf{64.63} \\
 & 768 & 88.01 & 87.49 & 88.24 & \textbf{88.38} & 89.15 & 88.43 & 89.14 & \textbf{89.61} & 
 & 768 & 62.5 & 63 & 60.14 & \textbf{63.07} & \textbf{64.78} & 64.21 & 63 & 64.35 \\
\midrule 

\multirow{7}{*}{TweetEval} 
 & 16 & 35.12 & 53.66 & 58.72 & \textbf{62.13} & 48.85 & 55.96 & 47.27 & \textbf{56.13} & \multirow{7}{*}{SICK} 
 & 16 & 60.66 & 59.07 & 62.56 & \textbf{64.3} & 45.75 & 59.35 & 61.5 & \textbf{65.09} \\
 & 32 & 46.06 & 58.74 & 62.78 & \textbf{64.79} & 55.5 & 60.51 & 54.15 & \textbf{57.48} & 
 & 32 & 64.69 & 65.08 & 64.98 & \textbf{67} & 51.86 & 63.34 & 62.36 & \textbf{67.08} \\
 & 64 & 49.72 & 60.22 & 64.96 & \textbf{68} & \textbf{63.27} & 58.39 & 61.09 & 62.09 & 
 & 64 & 67.5 & 68.17 & 66.92 & \textbf{68.82} & 52.97 & 66.11 & 62.07 & \textbf{69.09} \\
 & 128 & 55.89 & 66.71 & 66.75 & \textbf{69.01} & 63.92 & 60.52 & 65.24 & \textbf{65.63} & 
 & 128 & 68.91 & 68.88 & 66.62 & \textbf{69.66} & 59.33 & 66.57 & 66.74 & \textbf{70.29} \\
 & 256 & 63.54 & 67.88 & 68.97 & \textbf{71.46} & 65.77 & 63.05 & 66.62 & \textbf{68.48} & 
 & 256 & 68.93 & 68.85 & 67 & \textbf{69.92} & 65.72 & 67.23 & 66.43 & \textbf{70.51} \\
 & 512 & 66.47 & 68.01 & 69.32 & \textbf{72.36} & 67.46 & 65.72 & 67.78 & \textbf{69.73} & 
 & 512 & 69.87 & 69.47 & 67.69 & \textbf{70.13} & 67.41 & 68.33 & 67.04 & \textbf{70.84} \\
 & 768 & 70.16 & 69.32 & 70.02 & \textbf{72.69} & 68.08 & 66.6 & 69.05 & \textbf{69.34} & 
 & 768 & 69.93 & 69.53 & 67.76 & \textbf{70.14} & 68.15 & 68.94 & 68.19 & \textbf{70.75} \\
\midrule

\multirow{7}{*}{MRPC} 
 & 16 & 51.32 & 61.12 & 71.71 & \textbf{72.7} & 57.83 & 65.93 & 66.73 & \textbf{73.04} & \multirow{7}{*}{STSB} 
 & 16 & 55.88 & 57.84 & 57.45 & \textbf{58.82} & 49.24 & 61.65 & 59.39 & \textbf{63.36} \\
 & 32 & 57.34 & 69.87 & 72.28 & \textbf{73.27} & 66.72 & 70.11 & 71.07 & \textbf{73.56} & 
 & 32 & 62.16 & 62.97 & 60.12 & \textbf{63.67} & 51.34 & 65.09 & 64.39 & \textbf{66.8} \\
 & 64 & 68.25 & 71.44 & 72.98 & \textbf{73.33} & 70.53 & 70.14 & 72.57 & \textbf{73.39} & 
 & 64 & 58.37 & 65.45 & 65.78 & \textbf{65.93} & 61.22 & 67.32 & 67.51 & \textbf{69.02} \\
 & 128 & 68.15 & 72.07 & 72.52 & \textbf{73.45} & 73.14 & 72.96 & 71.71 & \textbf{73.73} & 
 & 128 & 65.11 & 67.16 & 65.61 & \textbf{67.46} & 67.53 & 69.59 & 70.12 & \textbf{71.04} \\
 & 256 & 71.82 & 72.31 & 72.63 & \textbf{73.69} & 73.07 & 72.98 & 72.29 & \textbf{73.41} & 
 & 256 & 67.24 & 67.79 & 65.01 & \textbf{68.23} & 68.23 & 70.41 & 70.91 & \textbf{72.23} \\
 & 512 & 71.97 & 72.6 & 72 & \textbf{73.74} & 73.79 & 73.16 & 73.27 & \textbf{73.82} & 
 & 512 & 67.82 & 68.85 & 68.92 & \textbf{69.12} & 70.67 & 71.33 & 70.83 & \textbf{73.2} \\
 & 768 & 72.18 & 72.85 & 72.01 & \textbf{73.91} & 73.91 & 73.22 & 72.46 & \textbf{73.96} & 
 & 768 & 69.31 & 68.93 & 69.02 & \textbf{69.38} & 71.58 & 71.3 & 71.99 & \textbf{72.88} \\
\bottomrule
\end{tabular}%
}
\caption{Results on in-domain datasets with TinyBERT 6L and BERT backbone embedding models. Bold indicates the best result at the same representation size, backbone model, and dataset.}
\label{tab1}
\end{table*}

\begin{table*}[h!]
\centering
\resizebox{\textwidth}{!}{%
\begin{tabular}{lc|cccc|cccc!{\vrule width 1.5pt}lc|cccc|cccc}
\toprule
\multirow{2}{*}{\textbf{Datasets}} & \multirow{2}{*}{\textbf{Dim.}} & \multicolumn{4}{c|}{\textbf{TinyBERT 6L}} & \multicolumn{4}{c!{\vrule width 1.5pt}}{\textbf{BERT}} & 
\multirow{2}{*}{\textbf{Datasets}} & \multirow{2}{*}{\textbf{Dim.}} & \multicolumn{4}{c|}{\textbf{TinyBERT 6L}} & \multicolumn{4}{c}{\textbf{BERT}} \\
\cmidrule(lr){3-6} \cmidrule(lr){7-10} \cmidrule(lr){13-16} \cmidrule(lr){17-20}
 & & \small{Unsup SimCSE} & \small{MRL} & \small{ESE} & \small{MIC} & \small{Unsup SimCSE} & \small{MRL} & \small{ESE} & \small{MIC} & 
 & & \small{Unsup SimCSE} & \small{MRL} & \small{ESE} & \small{MIC} & \small{Unsup SimCSE} & \small{MRL} & \small{ESE} & \small{MIC} \\
\midrule

\multirow{7}{*}{STS12} 
 & 16 & 45.68 & 51.48 & 50.9 & \textbf{57} & 47.88 & 55.13 & 51.34 & \textbf{60.86} & \multirow{7}{*}{STS16} 
 & 16 & 55.47 & 58.21 & 59.02 & \textbf{60.64} & 50.78 & 54.78 & 59.67 & \textbf{63.76} \\
 & 32 & 47.09 & 55.25 & 56.02 & \textbf{61.64} & 53.84 & 59.78 & 53.78 & \textbf{63.76} & 
 & 32 & 62.33 & 63.23 & 63.78 & \textbf{64.53} & 53.34 & 60.96 & 64.54 & \textbf{67.2} \\
 & 64 & 53.48 & 55.67 & 58.32 & \textbf{63.36} & 61.22 & 61.24 & 55.23 & \textbf{64.95} & 
 & 64 & 65.96 & 66.22 & 65.64 & \textbf{67.01} & 57.23 & 63.45 & 68.15 & \textbf{69.82} \\
 & 128 & 59.5 & 59.8 & 60.38 & \textbf{64.72} & 65.41 & 62.34 & 59.97 & \textbf{66.39} & 
 & 128 & 67.77 & 67.9 & 66.87 & \textbf{68.52} & 65.89 & 66.78 & 70.14 & \textbf{71.51} \\
 & 256 & 61.14 & 60.17 & 61.23 & \textbf{64.95} & 65.83 & 64.13 & 60.33 & \textbf{67.07} & 
 & 256 & 67.25 & 68.1 & 66.08 & \textbf{68.7} & 66.86 & 69.89 & 70.82 & \textbf{72.32} \\
 & 512 & 61.77 & 61.52 & 60.82 & \textbf{65.16} & 66.16 & 65.09 & 61.06 & \textbf{67.9} & 
 & 512 & 68.94 & 68.6 & 67.21 & \textbf{69.18} & 70.25 & 70.6 & 71.47 & \textbf{72.46} \\
 & 768 & 62.19 & 61.78 & 63.08 & \textbf{65.73} & 66.31 & 64.84 & 60.43 & \textbf{67.54} & 
 & 768 & 69.26 & 68.69 & 67.81 & \textbf{69.47} & 71.07 & 70.53 & 70.54 & \textbf{72.18} \\
\midrule

\multirow{7}{*}{STS13} 
 & 16 & 49.58 & 55.03 & 53.51 & \textbf{63.81} & 55.34 & 62.13 & 60.48 & \textbf{65.31} & \multirow{7}{*}{SickR} 
 & 16 & 60.41 & 63.88 & 60.79 & \textbf{64.42} & 55.67 & 61.92 & 60.92 & \textbf{65.31} \\
 & 32 & 50.6 & 63.48 & 60.46 & \textbf{68.25} & 61.23 & 63.44 & 65.78 & \textbf{68.97} & 
 & 32 & 64.41 & 66.29 & 64.69 & \textbf{66.9} & 57.23 & 64.75 & 63.96 & \textbf{67.48} \\
 & 64 & 58.43 & 62.96 & 64.56 & \textbf{69.39} & 67.73 & 68.98 & 68.97 & \textbf{72.27} & 
 & 64 & 67.2 & 67.84 & 65.7 & \textbf{68.63} & 60.86 & 66.92 & 65.63 & \textbf{69.34} \\
 & 128 & 60.93 & 67.88 & 66.4 & \textbf{70} & 71.09 & 70.97 & 71.68 & \textbf{74.24} & 
 & 128 & 68.78 & 68.72 & 66.47 & \textbf{69.44} & 64.23 & 67.45 & 67.19 & \textbf{70.72} \\
 & 256 & 61.24 & 67.8 & 68.53 & \textbf{70.62} & 71.84 & 71.24 & 72.32 & \textbf{75.37} & 
 & 256 & 69.12 & 69.18 & 66.91 & \textbf{70.73} & 66.78 & 68.04 & 67.63 & \textbf{70.55} \\
 & 512 & 67.92 & 68.96 & 69.65 & \textbf{71.42} & 72.13 & 72.35 & 73.63 & \textbf{76.02} & 
 & 512 & 69.75 & 69.43 & 67.74 & \textbf{70.01} & 67.34 & 68.56 & 68.61 & \textbf{70.87} \\
 & 768 & 70.54 & 69.16 & 69.94 & \textbf{71.59} & 74.71 & 73.56 & 73.81 & \textbf{76.52} & 
 & 768 & 69.82 & 69.49 & 67.82 & \textbf{70.32} & 67.89 & 68.76 & 68.14 & \textbf{70.81} \\
\midrule

\multirow{7}{*}{STS14} 
 & 16 & 49.35 & 50.42 & 52.23 & \textbf{55.61} & 50.67 & 51.76 & 54.2 & \textbf{56.08} & \multirow{7}{*}{Emotion} 
 & 16 & 28.44 & 29.47 & 29.76 & \textbf{31.51} & 24.78 & 26.98 & 23.92 & \textbf{29.66} \\
 & 32 & 48.23 & 53.67 & 55.98 & \textbf{59.15} & 57.23 & 54.78 & 56.89 & \textbf{59.71} & 
 & 32 & 32.65 & 36.77 & 36.11 & \textbf{38.91} & 34.15 & \textbf{36.87} & 31.4 & 34.75 \\
 & 64 & 51.28 & 59.57 & 59.25 & \textbf{60.96} & 60.86 & 54.78 & 60.44 & \textbf{62.36} & 
 & 64 & 41.35 & 42.35 & 44.29 & \textbf{46.83} & 42.46 & 42.52 & 38.02 & \textbf{46.63} \\
 & 128 & 55.98 & 60.84 & 60.72 & \textbf{62.37} & 61.25 & 62.05 & 61.45 & \textbf{63.97} & 
 & 128 & 46.39 & 49.5 & 52.16 & \textbf{52.19} & 49.76 & 49.15 & 45.76 & \textbf{52.3} \\
 & 256 & 60.13 & 60.91 & 62.36 & \textbf{63.06} & 61.56 & 63.68 & 63.81 & \textbf{65.75} & 
 & 256 & 52.53 & 50.69 & 56.12 & \textbf{56.87} & 54.42 & \textbf{54.67} & 48.63 & 54.72 \\
 & 512 & \textbf{63.93} & 62.22 & 61.85 & 63.59 & 62.45 & 64.56 & 64.63 & \textbf{66.91} & 
 & 512 & 54.93 & 55.01 & 56.82 & \textbf{59.48} & 54.36 & 55.98 & 51.39 & \textbf{58.83} \\
 & 768 & 63.01 & 62.32 & 62.23 & \textbf{63.84} & 65.58 & 64.65 & 64.85 & \textbf{66.93} & 
 & 768 & 55.3 & 57.44 & 60.43 & \textbf{61.31} & 53.78 & 54.08 & 51.17 & \textbf{60.36} \\
\midrule

\multirow{7}{*}{STS15} 
 & 16 & 62.85 & 64.17 & 63.84 & \textbf{65.47} & 61.45 & 61.99 & 60.29 & \textbf{66.92} & \multirow{7}{*}{SciTail} 
 & 16 & 71.26 & 71.35 & 70.69 & \textbf{71.78} & 68.15 & 67.45 & 69.14 & \textbf{73.09} \\
 & 32 & 70.06 & 69.88 & 69.23 & \textbf{70.86} & 65.34 & 69.86 & 65.87 & \textbf{70.07} & 
 & 32 & 72.38 & 72.24 & 72.06 & \textbf{72.76} & 69.33 & 67.26 & 69.71 & \textbf{74.02} \\
 & 64 & 69.32 & 72.51 & 72.08 & \textbf{73.59} & 65.78 & 72.08 & 67.57 & \textbf{72.68} & 
 & 64 & 72.71 & 72.48 & 72.1 & \textbf{74.17} & 69.56 & 67.21 & 70.1 & \textbf{74.71} \\
 & 128 & 70.54 & 73.24 & 73.75 & \textbf{74.67} & 66.23 & \textbf{75.28} & 69.86 & 74.89 & 
 & 128 & 71.77 & 72.06 & 72.53 & \textbf{74.23} & 70.46 & 67.73 & 71.69 & \textbf{74.65} \\
 & 256 & 74.76 & 73.29 & 72.29 & \textbf{75.56} & 71.32 & \textbf{76.24} & 70.67 & 75.97 & 
 & 256 & 71.35 & 71.58 & 71.77 & \textbf{74.18} & 70.79 & 67.82 & 71.42 & \textbf{74.27} \\
 & 512 & 75.48 & 73.87 & 75.24 & \textbf{75.97} & 74.56 & 76.19 & 71.58 & \textbf{76.48} & 
 & 512 & 71.45 & 71.77 & 72.15 & \textbf{74.74} & 70.64 & 67.92 & 71.46 & \textbf{74.5} \\
 & 768 & 75.75 & 73.9 & 76.19 & \textbf{76.23} & \textbf{77.08} & 76.62 & 70.14 & 76.62 & 
 & 768 & 71.2 & 71.78 & 72.38 & \textbf{74.5} & 70.55 & 67.63 & 71.88 & \textbf{74.18} \\
\bottomrule
\end{tabular}%
}
\caption{Results on out-domain datasets with TinyBERT 6L and BERT backbone embedding models. Bold indicates the best result at the same representation size, backbone model, and dataset.}
\label{tab2}
\end{table*}

Our framework enhances Matryoshka Representation Learning by combining a nested contrastive loss with two regularizers: Soft Collapse and Spectral Isotropy. This joint self-distillation strategy prevents redundancy and representation collapse.

\subsection{Soft Collapse Regularization}
Previous self-supervised methods such as Barlow Twins~\cite{zbontar2021barlowtwinsselfsupervisedlearning} emphasize reducing feature redundancy via cross-correlation constraints, but focus on global decorrelation rather than the structured dependency between nested subspaces. In MRL, this often leads to redundant features across prefix and residual dimensions, while enforcing hard orthogonality can be overly restrictive. To balance these effects, we introduce SCR module.

Let $\mathbf{H} \in \mathbb{R}^{B \times L \times d_{\text{full}}}$ denote the hidden state tensor and let $\mathbf{M} \in \{0,1\}^{B \times L}$ be the associated sequence attention mask. For a given truncation dimension $d < d_{\text{full}}$, we define the residual dimension $d_{\text{res}} = d_{\text{full}} - d$. The tensor is orthogonally partitioned into a prefix subspace $\mathbf{H}_{\text{pre}} \in \mathbb{R}^{B \times L \times d}$ and a residual subspace $\mathbf{H}_{\text{res}} \in \mathbb{R}^{B \times L \times d_{\text{res}}}$, such that $\mathbf{H} = [\mathbf{H}_{\text{pre}} \parallel \mathbf{H}_{\text{res}}]$.

To isolate geometric relationships from feature magnitudes, we perform sequence-wise standardization. Let $N_i = \sum_{l=1}^{L} M_{i,l}$ denote the active sequence length for the $i$-th batch element. The masked empirical moments for the prefix subspace are defined as:
\begin{equation}
\begin{aligned}
\boldsymbol{\mu}_{\text{pre}, i} &= \frac{1}{N_i} \sum_{l=1}^{L} M_{i,l} \mathbf{H}_{\text{pre}, i, l}, \\
\boldsymbol{\sigma}_{\text{pre}, i}^2 &= \frac{1}{N_i} \sum_{l=1}^{L} M_{i,l} \big(\mathbf{H}_{\text{pre}, i, l} - \boldsymbol{\mu}_{\text{pre}, i}\big)^{\odot 2}
\end{aligned}
\end{equation}
where $\odot 2$ denotes element-wise squaring. The standardized prefix tensor $\tilde{\mathbf{X}}_{\text{pre}} \in \mathbb{R}^{B \times L \times d}$ is computed point-wise:
$$\tilde{\mathbf{X}}_{\text{pre}, i, l} = M_{i,l} \left( \mathbf{H}_{\text{pre}, i, l} - \boldsymbol{\mu}_{\text{pre}, i} \right) \oslash \big( \boldsymbol{\sigma}_{\text{pre}, i} + \epsilon \big)$$
with $\oslash$ denoting Hadamard element-wise division. The standardized residual tensor $\tilde{\mathbf{X}}_{\text{res}} \in \mathbb{R}^{B \times L \times d_{\text{res}}}$ is derived symmetrically. The structural overlap between the two subspaces is quantified by the expected token-wise cross-correlation matrix $\mathbf{C} \in \mathbb{R}^{d \times d_{\text{res}}}$:
$$\mathbf{C} = \frac{1}{B} \sum_{i=1}^{B} \frac{1}{N_i} \sum_{l=1}^{L} \tilde{\mathbf{X}}_{\text{pre}, i, l} \tilde{\mathbf{X}}_{\text{res}, i, l}^\top$$

To mitigate subspace redundancy without imposing strict orthogonal constraints ($\mathbf{C} = \mathbf{0}$), we introduce a thresholded $\ell_2$ penalty over the correlation matrix. Let $\tau_{\text{corr}} \in [0, 1]$ act as the tolerance margin. The soft-collapse regularization loss is formulated as:
$$\mathcal{L}_{\text{corr}}^{(d)} = \frac{1}{d \cdot d_{\text{res}}} \sum_{u=1}^{d} \sum_{v=1}^{d_{\text{res}}} \max\big(0, |C_{u,v}| - \tau_{\text{corr}}\big)^2$$ 

To prevent dimensional collapse where the model minimizes correlation by simply shrinking the variance of certain dimensions to zero, we introduce a variance floor penalty. Let $\bar{\sigma}_{\text{pre}}$ and $\bar{\sigma}_{\text{res}}$ be the mean token-wise standard deviations of the respective subspaces. The variance loss is:
$$\mathcal{L}_{\text{var}}^{(d)} = \max(0, 1 - \bar{\sigma}_{\text{pre}}) + 0.5 \max(0, 1 - \bar{\sigma}_{\text{res}})$$
The lower weight of $0.5$ applied to the second term prioritizes the stability of the primary features while still providing enough signal to prevent the residual subspace from collapsing. This unbalanced weighting ensures that the optimization process remains focused on the most critical representative dimensions.
The total SCR loss for a given layer and dimension is a weighted sum:
$$\mathcal{L}_{\text{SCR}}^{(d)} =  \mathcal{L}_{\text{corr}}^{(d)} + \lambda_{\text{var}} \mathcal{L}_{\text{var}}^{(d)}$$

\subsection{Spectral Isotropy Regularization}
To improve the geometric properties of the representation space, we regularize the intermediate nested embeddings to be spectrally isotropic. Following principles of hyperspherical uniformity, we apply two sub-losses to the mean-pooled, nested representations $\mathbf{Z}^{(d)} \in \mathbb{R}^{B \times d}$.

First, we penalize the coefficient of variation (CV) of the per-dimension variance to ensure information is distributed evenly across all dimensions within the prefix. Let $\mathbf{Z}_{i}^{(d)} \in \mathbb{R}^d$ denote the mean-pooled prefix representation for the $i$-th sample in a batch of size $B$. The empirical variance for the $j$-th dimension across the batch is given by:
$$v_j = \frac{1}{B} \sum_{i=1}^{B} \left( Z_{i,j}^{(d)} - \mu_j \right)^2 \quad \text{where} \quad \mu_j = \frac{1}{B} \sum_{i=1}^{B} Z_{i,j}^{(d)}$$
We then define the mean $\bar{v}$ of these per-dimension variances over all $d$ dimensions $\bar{v} = \frac{1}{d} \sum_{j=1}^{d} v_j$. The coefficient of variation loss is formulated as the ratio of the standard deviation of the variances to their mean:

$$\mathcal{L}_{\text{cv}}^{(d)} = \frac{\sqrt{ \frac{1}{d} \sum_{j=1}^{d} (v_j - \bar{v})^2 }}{\bar{v} + \epsilon}$$

Second, to enforce a uniform distribution of the prefix embeddings upon the unit hypersphere $\mathcal{S}^{d-1}$, we minimize the empirical hyperspherical potential using a Radial Basis Function (RBF) kernel. Let $\hat{\mathbf{Z}}^{(d)} \in \mathbb{R}^{B \times d}$ be the row-normalized representation matrix for a batch of size $B$. We first compute the dense cosine similarity matrix $\mathbf{S} = \hat{\mathbf{Z}}^{(d)} (\hat{\mathbf{Z}}^{(d)})^\top$. Because the embeddings are unit-norm, the pairwise squared Euclidean distance between any two representations can be derived as: $\|\hat{\mathbf{z}}_i - \hat{\mathbf{z}}_j\|_2^2 = 2(1 - S_{ij})$. 

We construct the RBF kernel matrix $\mathbf{K} \in \mathbb{R}^{B \times B}$ where each element is defined as $K_{ij} = \exp\big(-2t(1 - S_{ij})\big)$. The uniformity loss seeks to minimize the average potential among all distinct pairs ($i \neq j$). We can express this average over the off-diagonal elements of the kernel matrix as:
$$\mathcal{L}_{\text{unif}}^{(d)} = \log \left( \frac{1}{B(B-1)} \Big( \mathbf{1}^\top \mathbf{K} \mathbf{1} - \text{Tr}(\mathbf{K}) \Big) + \epsilon \right)$$
where $\mathbf{1} \in \mathbb{R}^B$ is the vector of all ones, $\text{Tr}(\cdot)$ denotes the trace operator to exclude self-potentials, and $t = 2.0$ scales the kernel bandwidth. The combined spectral isotropy loss is
$\mathcal{L}_{\text{SIR}}^{(d)} = \frac{1}{2}(\mathcal{L}_{\text{cv}}^{(d)} +  \mathcal{L}_{\text{unif}}^{(d)})$. To further illustrate our findings, we include visualizations in Appendix~\ref{sec:visual}.
\subsection{Overall Training Objective}
The SCR and SIR regularizations are applied to a specific subset of intermediate transformer layers to enforce well-conditioned representations throughout the depth of the network. Detailed results for the intermediate layer selection experiments are documented in Appendix~\ref{sec:layers_select}. The total alignment loss across all selected layers and nested dimensions is the average of the regularization terms:

$$\mathcal{L}_{\text{align}} = \frac{1}{|L_{\text{align}}| |\mathcal{D}|} \sum_{l \in L_{\text{align}}} \sum_{d \in \mathcal{D}} \left( \mathcal{L}_{\text{SCR}}^{(l, d)} + \mathcal{L}_{\text{SIR}}^{(l, d)} \right)$$

 The final training objective combines the structural alignment regularization with the Matryoshka contrastive loss
$\mathcal{L}_{\text{total}} = \mathcal{L}_{\text{MRL}} + \gamma \mathcal{L}_{\text{align}}$.
Best hyper-parameters $\gamma$, $\tau_{\text{corr}}$ and $\lambda_{\text{var}}$ after the searching process are provided in Appendix~\ref{sec:hyper_params}. We also report additional experiments and loss components contributions in table~\ref{tab:ablation_study} from Appendix~\ref{sec:analysis} respectively.

\section{Experiments}

\subsection{Experimental Design}

To rigorously validate the representational capacity of the MIC framework, we design a comprehensive evaluation protocol spanning both in-domain (ID) and out-of-domain (OOD) scenarios.
\paragraph{Evaluation Benchmarks}
To test the global semantic representations in Text Classification task, we utilize TweetEval \citep{barbieri-etal-2020-tweeteval}, alongside Emotion and Banking77 from the MTEB suite \citep{muennighoff2023mtebmassivetextembedding}. We also evaluate how well the embedding subspaces capture relational semantics between text pairs in Natural Language Inference (NLI) task on MRPC \citep{wang-etal-2018-glue}, WiC \citep{pilehvar-camacho-collados-2019-wic}, and SciTail \citep{scitail}. About the Semantic Textual Similarity (STS) task, we conduct ID evaluations on STS-B and SICK \citep{marelli-etal-2014-semeval}, and extend to STS12-16 and SickR \citep{muennighoff2023mtebmassivetextembedding} to test OOD robustness. Comprehensive details regarding model architectures, hyperparameters, dataset statistics, and evaluation protocols are deferred to Appendix~\ref{sec:appendix_experiments}.

\paragraph{Baselines for Comparison} We benchmark MIC against methods in elastic-dimension representation learning: standard MRL~\cite{NEURIPS2022_c32319f4}, which serves as the foundational baseline for multi-granular contrastive training, and ESE~\cite{li2025ese}, a recent advancement that dynamically scales across both embedding width and network depth. Comprehensive metrics regarding training efficiency are provided in Appendix~\ref{sec:training_time}.

\subsection{Main Results}

Results in Table \ref{tab1} and \ref{tab2} show that MIC consistently matches or exceeds the performance of established baselines like Unsup SimCSE, MRL, and ESE. While maintaining parity in higher-dimensional regimes, MIC’s advantage becomes most pronounced at dimensions 16, 32, and 64, demonstrating superior semantic compression across tasks such as Banking77, TweetEval, and MRPC. This trend remains remarkably consistent across both TinyBERT and BERT architectures; for instance, at 16 dimensions on Banking77, MIC achieves scores of 44.1 and 59.45 respectively, significantly outperforming the next best alternatives. These findings validate the framework's robustness, confirming that MIC preserves semantic density and informational integrity under extreme truncation where traditional methods see a sharp decline. Furthermore, evaluations on BGE-M3 in Appendix~\ref{app:large_model} confirm that our subspace alignment strategy scales effectively to high-capacity encoders.

\section{Conclusion and Future Works}
In summary, we have introduced MIC, a framework using SCR and SIR to mitigate redundancy and dimensional collapse in Matryoshka Representation Learning. By optimizing for subspace independence and hyperspherical uniformity, MIC achieves superior semantic density even under extreme truncation. Future research will extend these geometric regularizers to generative and multi-modal settings while investigating automated methods for dynamic dimension selection.

\section{Limitations}
While our framework enhances semantic density, it introduces a trade-off in architectural flexibility: the layer-wise alignment requires a fixed mapping between specific Transformer layers and nested dimensions. This dependency makes performance gains sensitive to the backbone’s depth and configuration, potentially necessitating re-calibration for different architectures. Additionally, while SIR effectively promotes hyperspherical uniformity, the current implementation applies equal weighting to all nested dimensions during alignment, rather than dynamically scaling importance.

\section*{Acknowledgement}
This work was supported by VinUniversity’s Seed Grant Program under project VUNI.2425.EME.005.
\nocite{langley00}

\bibliography{example_paper}
\bibliographystyle{icml2026}

\newpage
\appendix
\onecolumn

\section{Related Works and Background}
\label{sec:rel_works}
\subsection{Related Works}
The trajectory of sentence representation learning has transitioned from aggregating static lexical vectors \citep{mikolov2013distributedrepresentationswordsphrases, pennington-etal-2014-glove} to the adoption of deep contextualized encoders \citep{peters-etal-2018-deep, devlin2019bertpretrainingdeepbidirectional}. While initial Transformer-based models produced anisotropic narrow embedding cones, Sentence-BERT \citep{reimers2019sentencebertsentenceembeddingsusing} pioneered the use of Siamese architectures to derive semantically meaningful sentence-level pools. To further refine the geometry of these spaces, contrastive frameworks such as SimCSE \citep{gao2022simcsesimplecontrastivelearning} and EASE \citep{nishikawa-etal-2022-ease} utilized dropout-based augmentation to enforce better feature distribution. Most recently, the field has pivoted toward leveraging the latent knowledge of Large Language Models (LLMs) to generate high-fidelity embeddings through instruction tuning and architectural adaptations \citep{llm2vec, he2025refiningsentenceembeddingmodel}. Despite these qualitative gains, the resulting high-dimensional vectors remain computationally expensive for large-scale retrieval, motivating a shift toward more flexible, dimension-adaptive architectures.

Matryoshka Representation Learning (MRL) \citep{NEURIPS2022_c32319f4} introduced a paradigm shift by nesting lower-dimensional sub-vectors within a high-dimensional representation, allowing for seamless inference-time truncation without the need for multiple specialized models. By supervising only a logarithmic number of dimensions, MRL facilitates an efficient coarse-to-fine search process. Recent extensions, such as Espresso Sentence Embeddings (ESE) \citep{li2025ese}, have introduced compress-and-express modules to bridge the performance gap across varying depths and widths. The Matryoshka principle has proven versatile beyond text, seeing successful integration into diffusion-based image generation \citep{gu2024matryoshkadiffusionmodels}, multimodal alignment \citep{cai2024matryoshkamultimodalmodels}, and Query Transformer architectures for vision-language tasks \citep{hu2024matryoshkaquerytransformerlarge}. However, existing research has largely focused on the functional usability of nested dimensions through multi-objective loss functions. Our work, MIC, diverges from these efforts by focusing on the internal geometry, specifically aiming to solve the problems of subspace redundancy and spectral collapse that often plague highly compressed Matryoshka prefixes.

\subsection{Background}

\subsubsection{Matryoshka Representation Learning}

Let $f_\theta: \mathcal{X} \to \mathbb{R}^{d_{\text{max}}}$ denote a parameterized neural encoder that maps a discrete text input $x \in \mathcal{X}$ to a dense, continuous vector representation $\mathbf{h} = f_\theta(x)$. Conventional representation learning optimizes $\mathbf{h}$ exclusively at its maximum dimensionality $d_{\text{max}}$. In MRL~\cite{NEURIPS2022_c32319f4}, the goal is to optimize a single embedding such that multiple nested sub-vectors prefixes are simultaneously highly expressive.

Formally, we define an ordered set of truncation dimensions $\mathcal{M} = \{m_1, m_2, \dots, m_k\}$, where $0 < m_1 < m_2 < \dots < m_k = d_{\text{max}}$. We define a deterministic truncation operator $\Pi_m: \mathbb{R}^{d_{\text{max}}} \to \mathbb{R}^m$, which acts as an orthogonal projection matrix $\mathbf{P}_m = [\mathbf{I}_m \;\; \mathbf{0}_{m \times (d_{\text{max}} - m)}]$ onto the first $m$ basis vectors:
$$\mathbf{h}^{(m)} = \Pi_m(\mathbf{h}) = \mathbf{P}_m \mathbf{h}$$

In the context of contrastive learning, which serves as the primary optimization mechanism for modern dense retrievers, MRL extends the standard InfoNCE objective across all granularities $m \in \mathcal{M}$. Given a batch of positive sentence pairs $\mathcal{B} = \{(x_i, x_i^+)\}_{i=1}^B$, the Matryoshka contrastive loss uniformly supervises each nested subspace:

\begin{equation}
\begin{aligned}
\mathcal{L}_{\text{MRL}}(\theta) = - \frac{1}{|\mathcal{M}| B} \sum_{m \in \mathcal{M}} \sum_{i=1}^{B} \bigg[ &\frac{\text{sim}(\mathbf{h}_i^{(m)}, \mathbf{h}_i^{+(m)})}{\tau} 
- \log \sum_{j=1}^{B} \exp\bigg( \frac{\text{sim}(\mathbf{h}_i^{(m)}, \mathbf{h}_j^{+(m)})}{\tau} \bigg) \bigg]
\end{aligned}
\end{equation}

where $\text{sim}(\mathbf{u}, \mathbf{v}) = \frac{\mathbf{u}^\top \mathbf{v}}{\|\mathbf{u}\|_2 \|\mathbf{v}\|_2}$ is the cosine similarity and $\tau$ is a temperature scalar. While Equation 2 ensures that early dimensions capture task-relevant semantics, it does not constrain the internal covariance structure or spectral distribution of the subspaces, often leading to severe capacity degradation at lower dimensions.

\subsubsection{Representation Geometry: Isotropy and Subspace Redundancy}

To understand the failure modes of highly compressed Matryoshka prefixes, we must analyze the geometric properties of the embedding space. Ideal contrastive representations should exhibit hyper-spherical uniformity~\cite{pmlr-v119-wang20k}, meaning the normalized feature vectors are evenly distributed over the unit hyper-sphere $\mathcal{S}^{d-1} = \{ \mathbf{v} \in \mathbb{R}^d : \|\mathbf{v}\|_2 = 1 \}$. 

The uniformity of a continuous distribution $p_{\text{data}}$ over $\mathcal{S}^{d-1}$ can be quantified by examining the expected pairwise distances under a Gaussian potential with Radial Basis Function kernel:
$$\mathcal{L}_{\text{uniform}}(f; t) = \log \mathbb{E}_{x, y \sim p_{\text{data}}} \left[ \exp \left( -t \big\| \tilde{f}(x) - \tilde{f}(y) \big\|_2^2 \right) \right]$$
where $\tilde{f}(\cdot)$ denotes the $\ell_2$-normalized output and $t > 0$ controls the kernel bandwidth. Transformer-based models inherently suffer from the anisotropy problem~\cite{ethayarajh-2019-contextual}, where embeddings degenerate into a narrow cone, causing $\mathcal{L}_{\text{uniform}}$ to be sub-optimally high.

Furthermore, within the Matryoshka framework, information capacity is fundamentally bounded by the spectral rank of the empirical covariance matrix. Let $\mathbf{\Sigma} \in \mathbb{R}^{d_{\text{max}} \times d_{\text{max}}}$ be the covariance matrix of the full representations over a batch. For a given truncation threshold $m$, $\mathbf{\Sigma}$ can be partitioned into block matrices corresponding to the prefix subspace and the residual subspace:
$$ \mathbf{\Sigma} = \begin{bmatrix} 
\mathbf{\Sigma}_{\text{pre}} & \mathbf{\Sigma}_{\text{cross}} \\ 
\mathbf{\Sigma}_{\text{cross}}^\top & \mathbf{\Sigma}_{\text{res}} 
\end{bmatrix} $$
where $\mathbf{\Sigma}_{\text{pre}} \in \mathbb{R}^{m \times m}$ is the covariance of the prefix, and $\mathbf{\Sigma}_{\text{cross}} \in \mathbb{R}^{m \times (d_{\text{max}} - m)}$ is the cross-covariance between the prefix and the residual dimensions. In standard MRL, there are no geometric constraints placed on $\mathbf{\Sigma}_{\text{cross}}$. If the features in the prefix are highly correlated with the residual features, the entries of $\mathbf{\Sigma}_{\text{cross}}$ are non-zero, indicating overlapping linear dependencies. This redundancy implies that the effective dimensionality and the Shannon informational capacity of the concatenated vector is strictly less than its algebraic dimension. Additionally, if the eigenvalues of $\mathbf{\Sigma}_{\text{pre}}$ decay rapidly, the coefficient of variation of the per-dimension variance increases, resulting in a fragile subspace where only a few leading dimensions govern the similarity metric.

\section{Experimental Details}
\label{sec:appendix_experiments}

\subsection{Model Architectures}
To evaluate the scalability and generalizability of the MIC framework, we conduct experiments across a diverse range of backbone architectures varying in size and complexity. These include the compact \textbf{TinyBERT-6L} with 6 Transformer layers, the standard 12-layer \textbf{BERT-base} encoder, and the large-scale \textbf{BGE-M3} model. To ensure robustness of our experiments, we train each model in three independent runs with different seeds and report average score. By spanning across language embedding models, we demonstrate that MIC consistently optimizes the internal organization of semantic information.

\subsection{Detailed Dataset Statistics}

\begin{table*}[htbp]
\centering
\caption{Dataset Statistics for training and test set}
\label{tab:dataset-statistics}
\resizebox{0.5\textwidth}{!}{%
\begin{tabular}{lcc}
\toprule
\textbf{Dataset} & \textbf{Train (Sampled)} & \textbf{Test Size} \\
\midrule
Banking77 & 3,000 & 3,080 \\
TweetEval & 3,000 & 3434  \\
Emotion (OOD) & - & 1,990  \\
MRPC & 1,500 & 1,730  \\
WiC & 1,500 & 1,400 \\
SciTail (OOD) & - & 2,130 \\
SICK & 1,500 & 4,823 \\
STS-B & 1,500 & 1,390 \\
STS12 (OOD) & - & 3,108 \\
STS13 (OOD) & - & 1,500 \\
STS14 (OOD) & - & 3,750 \\
STS15 (OOD) & - & 3,000 \\
STS16 (OOD) & - & 1,186 \\
SickR (OOD) & - & 9,927 \\
\bottomrule
\end{tabular}}
\end{table*}
We use a diverse collection of datasets for both training and evaluation, covering text classification, natural language inference (NLI), and semantic textual similarity (STS). To build training data for contrastive sentence representation learning, we sample sentences from multiple task categories to promote domain and objective diversity. Specifically, we collect 6,000 sentences from classification datasets (3,000 per dataset), 3,000 sentence pairs from STS datasets (1,500 per dataset), and 3,000 sentence pairs from pair classification datasets (1,500 per dataset). All sentence pairs are flattened into individual sentences, resulting in 24,000 unique sentences, which are then used to train the backbone encoder with unsupervised SimCSE-style contrastive learning under a unified training framework. For evaluation, we test the trained models on both in-domain test sets and unseen out-of-domain datasets, including Emotion, SciTail, and multiple STS benchmarks (STS12–STS16 and SickR), which differ from the training data in domain, style, and annotation protocol. Dataset statistics are reported in Table~\ref{tab:dataset-statistics}, and all baseline methods (MRL, ESE) are trained on the same training corpus and evaluated on the same set of test datasets for fair comparison

\subsection{Training Configurations}
\label{sec:hyper_params}
The detailed training configurations for the MIC framework across our various backbones are summarized in Table~\ref{tab:training-config}. We maintain a consistent setup to ensure a fair comparison across different scales, applying specific hyperparameters such as $\tau_{\text{corr}}$, $\gamma$ and $\lambda_{\text{var}}$ between loss components. We explored the loss balancing hyperparameter $\tau_{\text{corr}}$, $\gamma$ and $\lambda_{\text{var}}$ both over the set $\{0.1, 0.2, 0.3,0.4, 0.5,0.6, 0.7, 0.8, 0.9, 1.0\}$. The optimal configurations for each backbone are also reported in Table~\ref{tab:training-config}.

\begin{table*}[ht]
\centering
\caption{Detailed training configurations for MIC across different backbones.}
\label{tab:training-config}
\begin{tabular}{lccc}
\toprule
\textbf{Configuration} & \textbf{TinyBERT-6L} & \textbf{BERT-base} & \textbf{BGE-M3} \\
\midrule
Epochs & 5 & 5 & 5 \\
Learning Rate & $2 \times 10^{-5}$ & $2 \times 10^{-5}$ & $2 \times 10^{-5}$ \\
Max Length & 256 & 256 & 256 \\
Batch Size & 32 & 32 & 32 \\
LR Scheduler & Cosine & Cosine & Cosine \\
Optimizer & AdamW & AdamW & AdamW \\
$\gamma$ & 0.6 & 0.6 & 0.6 \\
$\lambda_{\text{var}}$ & 0.1 & 0.1 & 0.1 \\
$\tau_{\text{corr}}$ & 0.1 & 0.1 & 0.1 \\
\bottomrule
\end{tabular}
\end{table*}

\paragraph{Task Objective Specification.} In our experimental setting, the task-specific loss component within the MRL objective is instantiated as $\mathcal{L}_{\text{SimCSE}}$, the unsupervised contrastive loss adopted from \citep{gao2022simcsesimplecontrastivelearning} . Given a batch of input sentences, we feed them through the student encoder twice with different standard dropout masks $z, z'$ to obtain two views of embeddings $e_i^z$ and $e_i^{z'}$. The loss is formulated as:
\begin{equation}
    \mathcal{L}_{\text{SimCSE}} = -\frac{1}{N} \sum_{i=1}^{N} \log \frac{e^{\text{sim}(e_i^z, e_i^{z'}) / \tau}}{\sum_{j=1}^{N} e^{\text{sim}(e_i^z, e_j^{z'}) / \tau}},
    \label{eq:simcse}
\end{equation}
where $N$ is the batch size, $\tau$ is the temperature hyperparameter, and $\text{sim}(\cdot)$ denotes cosine similarity.

\begin{table}[htbp]
\centering
\small
\caption{Performance comparison on Banking77 and Emotion datasets using BERT.}
\label{tab:ablation_study}
\setlength{\tabcolsep}{8pt}
\begin{tabular}{l cccc}
\toprule
\textbf{Datasets} & \multicolumn{4}{c}{\textbf{BERT}} \\
\cmidrule{2-5}
 & \textbf{MRL} & \textbf{SCR} & \textbf{SIR} & \textbf{MIC} \\
\midrule

\multirow{7}{*}{Banking77}
& 46.39 & 49.56 & 48.96 & \textbf{59.45} \\
& 64.90 & 69.09 & 68.20 & \textbf{75.71} \\
& 76.84 & 79.44 & 79.47 & \textbf{83.05} \\
& 83.49 & 84.42 & 84.61 & \textbf{86.57} \\
& 86.45 & 87.11 & 87.35 & \textbf{88.11} \\
& 87.85 & 88.28 & 88.58 & \textbf{89.43} \\
& 88.43 & 89.22 & 88.97 & \textbf{89.61} \\
\midrule

\multirow{7}{*}{Emotion}
& 29.47 & 30.56 & 31.05 & \textbf{31.51} \\
& 36.77 & 37.82 & 37.51 & \textbf{38.91} \\
& 42.35 & 44.14 & 43.01 & \textbf{46.83} \\
& 49.50 & 48.30 & 47.60 & \textbf{52.19} \\
& 50.69 & 54.72 & 53.83 & \textbf{56.87} \\
& 55.01 & 57.69 & 56.28 & \textbf{59.48} \\
& 57.44 & 54.72 & 55.76 & \textbf{61.31} \\
\bottomrule

\end{tabular}
\end{table}

\subsection{Evaluation}

To quantify the representational utility of the learned Matryoshka prefixes, we evaluate our model across a spectrum of nested dimensions $\mathcal{D} = \{16, 32, 64, 128, 256, 512, 768/1024\}$ spanning three distinct downstream paradigms. For single-text classification, we adhere to the established frozen-feature protocol \citep{conneau2018sentevalevaluationtoolkituniversal} by training a Logistic Regression classifier on the fixed embeddings and reporting the F1 Score. In pair-wise classification tasks, such as NLI, we derive predictions by applying an optimal similarity threshold to the cosine distance between representations to report Accuracy, while for Semantic Textual Similarity (STS) benchmarks, we assess the alignment between model-derived similarities and human-annotated gold standards using the Spearman correlation coefficient. Throughout this process, all models are fully fine-tuned using our unified $\mathcal{L}_{\text{total}}$ objective, ensuring that the resulting performance metrics accurately reflect the structural and semantic density achieved through isotropic subspace alignment.

\section{Analysis}
\label{sec:analysis} 
\subsection{Importance of Loss Components}

\begin{figure*}[t]
  \centering
  \includegraphics[width=0.7\linewidth]{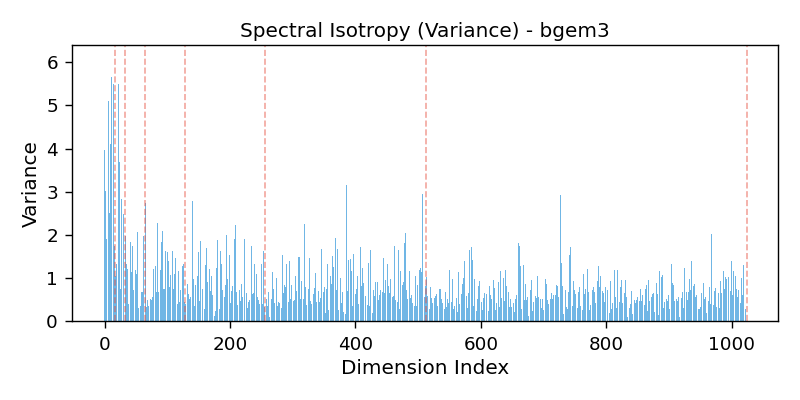} 
  \caption {Spectral Isotropy Analysis. Distribution of per-dimension variance across the embedding space. Our SIR framework prevents dimensional collapse by maintaining a balanced variance profile. This ensures that information is distributed across the entire vector rather than being concentrated in a few dominant dimensions.}
  \label{fig:bgem3_variance}
\end{figure*}

To further evaluate the robustness of our framework, we report the performance of MIC and its components on the Banking77 and Emotion datasets. As shown in the table~\ref{tab:ablation_study}, the fully integrated MIC framework consistently exceeds the performance of the standard MRL baseline across the entire Matryoshka hierarchy. The performance gap is particularly striking in extreme low-dimensional regimes; for instance, on the Banking77 dataset at the $d=16$ truncation point, MIC achieves a significant improvement of over 13 absolute points compared to MRL (59.45 vs. 46.39). This suggests that our geometric constraints successfully condense discriminative task signals into minimal subspaces that would otherwise suffer from informational sparsity. Furthermore, the ablation results for SCR and SIR confirm that while each module independently enhances representational fidelity over the baseline, their combined application in MIC yields a synergistic effect. This joint optimization ensures that the prefix dimensions are not only independent of the residual features but also uniformly distributed on the hypersphere, ultimately maximizing the informational capacity of the model without sacrificing performance as the dimensionality scales toward full capacity.

\subsection{Visualizations}
\label{sec:visual}
\begin{table}[h]
    \centering
    \caption{Training efficiency comparison on the BERT-base backbone. MIC incurs higher training latency due to the computation of alignment losses.}
    \resizebox{0.5\linewidth}{!}{
    \begin{tabular}{l|cc}
        \toprule
        \textbf{Method} & \textbf{Iterations / s} & \textbf{Throughput (sample/s)} \\
        \midrule
        SimSCE & 3.56 & 113.92 \\
        MRL & 6.70 & 214.4 \\
        ESE & 5.69 & 182.08 \\
        \textbf{MIC (Ours)} & 3.30 & 105.6 \\
        \bottomrule
    \end{tabular}
    }
    \label{tab:training_time}
\end{table}

Figure~\ref{fig:bgem3_variance} illustrates the per-dimension variance of the embedding space for the BGE-M3 model trained using our proposed method. The red dashed lines indicate the nested Matryoshka boundaries $d \in \{16, 32, 64, 128, 256, 512, 1024\}$. A common failure mode in embedding models is dimensional collapse, where the variance is concentrated in a tiny fraction of the available dimensions. In contrast, our SIR maintains a remarkably balanced variance profile across the entire 1024-dimensional vector. This isotropy is particularly evident within the smaller nested prefixes, ensuring that even highly compressed representations retain sufficient expressive power and information density to perform effectively on downstream tasks.

\begin{figure*}[t]
  \centering
  \includegraphics[width=0.6\linewidth]{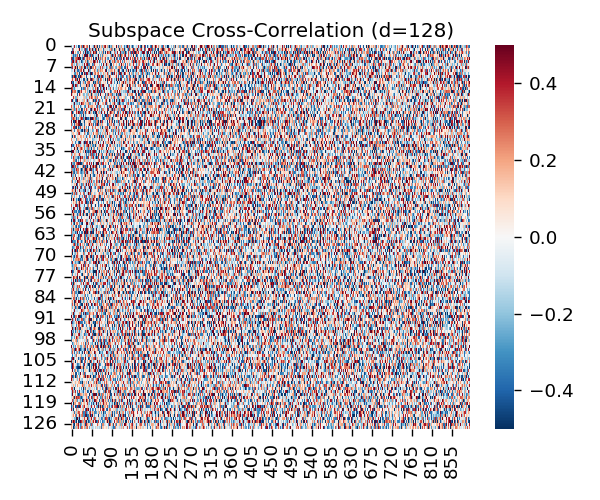} 

  \caption {Cross-correlation matrix between the $d=128$ prefix and residual subspaces. Near-zero values indicate successful de-correlation and non-redundant feature learning across nested dimensions.}
  \label{fig:bgem3_correlation}
\end{figure*}

Figure~\ref{fig:bgem3_correlation} visualizes the cross-correlation matrix between the $d=128$ prefix subspace and the residual dimensions of the full embedding. A key challenge in Matryoshka learning is ensuring that nested dimensions capture unique information rather than becoming redundant versions of one another. The heatmap reveals a predominantly neutral distribution with values concentrated near zero, demonstrating that our SCR module successfully de-correlates the prefix and residual subspaces. This lack of significant cross-correlation confirms that the model learns additive, non-redundant features as dimensionality increases, effectively preventing informational collapse and maximizing the representational efficiency of each nested tier.

\section{Layers Selection}
\label{sec:layers_select}
The selection of layers for each model is strategically guided by the empirical evidence shown in Table~\ref{tab:bert_mic_layers}, which demonstrates that relying solely on the "First Layer" or "Last Layer" often results in suboptimal performance. For BERT, layers 8 and 10 are chosen because late-intermediate layers in 12-layer models typically capture the most robust semantic information before the final layer becomes too specialized for the pre-training objective. This principle is applied to TinyBERT by selecting layers 2 and 4, which represent the corresponding mid-to-late stages of its shallower 6-layer architecture. Finally, for BGE-M3, a more distributed set of layers $\{4, 7, 11\}$ is utilized to capture a hierarchical progression of features across the model's depth.

Limiting the selection to 2-3 layers per model is a deliberate design choice to optimize the trade-off between model performance and training throughput. While aligning every layer might offer marginal gains in representation consistency, the computational cost and memory overhead scale linearly with the number of aligned layers. By targeting a sparse set of high-utility layers, the training process remains time-efficient and scalable without sacrificing the semantic richness required for high-quality Matryoshka embeddings.

\begin{table}[htbp]
\centering
\small
\caption{Performance comparison on Banking77 and MRPC datasets using BERT.}
\label{tab:bert_mic_layers}
\setlength{\tabcolsep}{8pt}
\begin{tabular}{l ccc}
\toprule
\textbf{Datasets} & \multicolumn{3}{c}{\textbf{BERT}} \\
\cmidrule{2-4}
 & \textbf{MIC} & \textbf{MIC (Last Layer)} & \textbf{MIC (First Layer)} \\
\midrule

\multirow{7}{*}{Banking77}
& \textbf{59.45} & 50.37 & 46.68 \\
& \textbf{75.71} & 68.20 & 64.98 \\
& \textbf{83.05} & 78.62 & 76.94 \\
& \textbf{86.57} & 84.65 & 84.13 \\
& \textbf{88.11} & 87.10 & 86.91 \\
& \textbf{89.43} & 88.19 & 88.91 \\
& \textbf{89.61} & 88.94 & 88.40 \\
\midrule

\multirow{7}{*}{MRPC}
& \textbf{73.04} & 71.19 & 71.48 \\
& \textbf{73.56} & 71.67 & 72.69 \\
& 73.39 & 72.81 & \textbf{73.50} \\
& \textbf{73.73} & 73.04 & 73.62 \\
& 73.41 & 73.39 & \textbf{73.68} \\
& \textbf{73.82} & 72.81 & 73.50 \\
& \textbf{73.96} & 73.10 & 73.44 \\
\bottomrule

\end{tabular}
\end{table}

\section{Training Time Analysis}
\label{sec:training_time}
We explicitly analyze the computational cost of our framework compared to the SimSCE, MRL and ESE baselines using the BERT-base backbone. The training throughput which is measured in iterations per second and samples per second is reported in Table \ref{tab:training_time}.

Table \ref{tab:training_time} provides a comparative analysis of training efficiency across the evaluated frameworks. While standard MRL achieves the highest throughput which is 214.4 samples/s due to its simple multi-head cross-entropy objective, MIC exhibits a training speed of 105.6 samples/s. This computational overhead is a direct consequence of the intensive geometric regularization performed during the self-distillation phase. Specifically, the SCR necessitates the computation of token-wise covariance and cross-correlation matrices across multiple layers and nested dimensions, while the SIR introduces $\mathcal{O}(B^2)$ pairwise distance calculations to enforce hyperspherical uniformity. Although MIC incurs a higher training latency compared to the SimSCE and ESE baselines, it is crucial to emphasize that these regularizers are applied exclusively during the optimization phase. At inference time, MIC retains the exact same architecture and latency profile as standard MRL, ensuring that the enhanced informational capacity and spectral properties of the embeddings are realized without any additional computational cost for downstream deployment.

\section{Larger Model Results}
\label{app:large_model}

The results in Table~\ref{tab:bgem3_results} present a comprehensive evaluation of BGEM3 across the Banking77 and SICK datasets under varying embedding dimensions, highlighting consistent performance trends among MRL, ESE, and MIC. Across both datasets, MIC consistently achieves the best performance at every dimensional setting, demonstrating its effectiveness in preserving semantic information during representation learning. This advantage becomes particularly evident at lower dimensions (e.g., 16 and 32), where the performance gap between MIC and the other methods is more pronounced, indicating its superior capability in handling aggressive compression. As the dimensionality increases, all methods show steady improvements, but MIC maintains a stable lead, suggesting better scalability and robustness. Additionally, the performance differences between methods gradually narrow at higher dimensions (e.g., 512 and 1024), implying that richer embedding spaces reduce the relative difficulty of the task. Overall, these findings confirm that MIC provides a more reliable and efficient encoding strategy across both low- and high-dimensional regimes, while also generalizing consistently across datasets with different characteristics.

\begin{table}[htbp]
\centering
\small
\caption{Performance comparison on Banking77 and SICK datasets using BGEM3.}
\label{tab:bgem3_results}
\setlength{\tabcolsep}{8pt}
\begin{tabular}{l cccc}
\toprule
\textbf{Datasets} & \textbf{Dim} & \multicolumn{3}{c}{\textbf{BGEM3}} \\
\cmidrule{3-5}
 &  & \textbf{MRL} & \textbf{ESE} & \textbf{MIC} \\
\midrule

\multirow{7}{*}{Banking77}
& 16   & 64,39 & 62,69 & \textbf{66,11} \\
& 32   & 76,42 & 73,7  & \textbf{78,17} \\
& 64   & 83,81 & 79,34 & \textbf{84,73} \\
& 128  & 86,39 & 83,92 & \textbf{87,05} \\
& 256  & 88,43 & 86,21 & \textbf{88,85} \\
& 512  & 88,69 & 87,48 & \textbf{89,05} \\
& 1024 & 88,97 & 88,03 & \textbf{89,48} \\
\midrule

\multirow{7}{*}{SICK}
& 16   & 61,99 & 61,3  & \textbf{62,64} \\
& 32   & 62,79 & 62,05 & \textbf{63,36} \\
& 64   & 63,33 & 62,58 & \textbf{63,79} \\
& 128  & 63,8  & 63,12 & \textbf{64,2} \\
& 256  & 64,18 & 63,52 & \textbf{64,6} \\
& 512  & 64,64 & 64,04 & \textbf{65,07} \\
& 1024 & 65,21 & 64,34 & \textbf{65,6} \\
\bottomrule

\end{tabular}
\end{table}

\end{document}